\documentclass[10pt,twocolumn,letterpaper]{article}

\usepackage[pagenumbers]{cvpr}

\usepackage{multirow}

\definecolor{cvprblue}{rgb}{0.21,0.49,0.74}
\usepackage[pagebackref,breaklinks,colorlinks,allcolors=cvprblue]{hyperref}

\newtcolorbox{PromptBlock}{
  enhanced,
  breakable,                
  colback=bg-gray,          
  colframe=light-gray,      
  boxrule=0.5pt,            
  arc=2pt,                  
  left=6pt, right=6pt, top=6pt, bottom=6pt,
  before skip=6pt, after skip=6pt,
  segmentation style={solid, dashed, draw=light-gray},    
}

\newcommand{\promptblock}[1]{%
  \begin{PromptBlock}\small
  \textcolor{black}{#1}
  \end{PromptBlock}%
}

\newcommand{\commenttext}[1]{\textcolor{light-gray}{#1}}

\definecolor{dark-gray}{HTML}{A9A9A9} %
\definecolor{light-gray}{HTML}{b7b7b7} %
\definecolor{light-green}{HTML}{dcf8c6}
\definecolor{light-blue}{HTML}{a5cdfb}
\definecolor{bg-gray}{HTML}{F8F8F8} 
	
\definecolor{dark-green}{HTML}{6aa84f} 
\definecolor{highlight}{HTML}{cfe2f3} 
\definecolor{blue}{HTML}{2B60DE} 
\definecolor{brown}{HTML}{9F8C76}
\definecolor{pink}{HTML}{D88782}
\definecolor{purple}{HTML}{9370DB}
\definecolor{dark-yellow}{HTML}{F6AE2D}
\definecolor{light-yellow}{HTML}{ebdc7d}

\def\paperID{*****} 
\def\confName{CVPR}
\def\confYear{2026}

\title{SoPE: Spherical Coordinate-Based Positional Embedding for Enhancing Spatial Perception of 3D LVLMs }

\author{
Guanting Ye$^{1}$ \quad
Qiyan Zhao$^{3}$ \quad
Wenhao Yu$^{2}$ \quad
Liangyu Yuan$^{4}$ \quad
Mingkai Li$^{5}$ \quad
Xiaofeng Zhang$^{3}$ \\
Jianmin Ji$^{2}$ \quad
Yanyong Zhang$^{2}$ \quad
Qing Jiang$^{4}$ \quad
Ka-Veng Yuen $^{1}$ \\
$^{1}$University of Macau \quad
$^{2}$University of Science and Technology of China \\  
$^{3}$Shanghai Jiaotong University \quad
$^{4}$Hefei University of Technology \quad
$^{5}$National University of Singapore \\
{\tt\small Corresponding author: kvyuen@um.edu.mo }
}

\begin{document}
\maketitle
\begin{abstract}
3D Large Vision-Language Models (3D LVLMs) built upon Large Language Models (LLMs) have achieved remarkable progress across various multimodal tasks. However, their inherited position-dependent modeling mechanism, Rotary Position Embedding (RoPE), remains suboptimal for 3D multimodal understanding. The vanilla RoPE formulation fails to preserve essential three-dimensional spatial structures when encoding 3D tokens, and its relative distance computation overlooks angular dependencies hindering the model’s ability to capture directional variations in visual representations. To overcome these limitations, we introduce Spherical Coordinate–based Positional Embedding (SoPE). Our method maps point-cloud token indices into a 3D spherical coordinate space, enabling unified modeling of spatial locations and directional angles. This formulation preserves the inherent geometric structure of point-cloud data, enhances spatial awareness, and yields more consistent and expressive geometric representations for multimodal learning. In addition, we introduce a multi-scale frequency mixing strategy to fuse feature information across different frequency domains. Experimental results on multiple 3D scene benchmarks validate the effectiveness of our approach, while real-world deployment experiments further demonstrate its strong generalization capability.
\end{abstract}

\section{Introduction}

Three-dimensional scenes are the physical substrate of everyday interaction, and enabling machines to perceive, reason about, and act within them is a central goal in AI. Recent efforts build 3D Large Vision-Language Models (LVLMs)~\cite{achiam2023gpt,chen2024ll3da,hu2025m,xue2025comfybench,zhu2024beyond,zheng2025video,zohar2025apollo,cheng2024spatialrgpt} that leverage pretrained LLMs to lift 3D scene understanding from geometric recognition toward richer semantic, functional, and intent-aware reasoning. To construct 3D visual features that can interface with language, prior work has followed several routes: projecting point clouds into images~\cite{hong20233d,zhi2024lscenellm,hao2024coarse,hess2024lidarclip,ding2024unireplknet,wang20243d}, directly encoding 3D geometry~\cite{chat-3d,xu2024pointllm,jin2025geoformer,bhattarai2024triplanenet,wu2024direct3d,tang2024any2point,zhang2025tar3d,zuo2024high}, or fusing multi-view images~\cite{hu2024mvd,xiao2024mim,ren2024pointofview,cheng2024adaptive,ren2024novel,gan2025multi} and point clouds as in LLaVA-3D~\cite{zhu2024llava}. These advances focus on the visual front end, but successful end-to-end 3D LVLMs also require connectors that preserve spatial and pose cues when aligning visual tokens to the LLM input space.

\begin{figure}[t]
\centering
\includegraphics[page=1,width=0.5\textwidth]{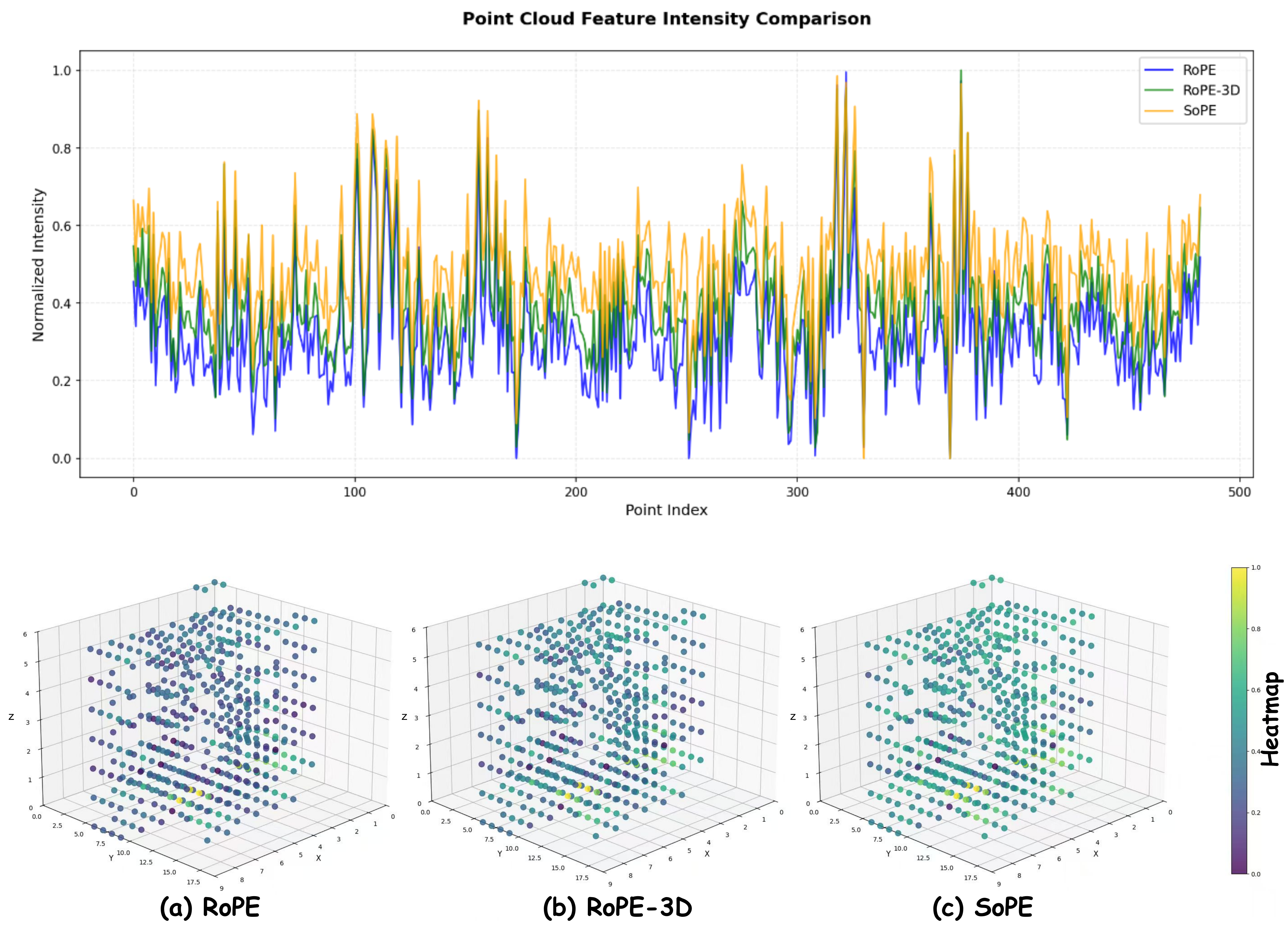} 
\caption{\textbf{ SoPE enhances 3D spatial representation via geometry-aware positional encoding.} Conventional positional encoding methods such as (a) RoPE and (b) RoPE-3D exhibit spatial perception bias: cross-modal attention concentrates on a few hotspots and fails to capture global context, resulting in less discriminative spatial features. In contrast, the proposed (c) SoPE produces more balanced and global attention patterns while amplifying spatial feature saliency, indicating a deeper understanding of the 3D environment.}
\label{fig1}
\end{figure} 

Unlike the most prior work, we take a positional-encoding perspective and focus on rotary positional embedding (RoPE)~\cite{su2024roformer,heo2024rotary}, which is widely adopted for position-dependent modeling in current 3D LVLMs. RoPE flattens point-cloud tokens into a 1D sequence in raster order and assigns positional indices according to their sequence positions, modeling positional dependencies via relative index differences. While this design is effective in NLP, in 3D settings it relies solely on 1D sequence indices to implicitly represent geometry and thus ignores the inherent 3D attributes of point-cloud tokens, such as their true spatial locations and orientations. This mismatch between the index-based positional modeling and the underlying point-cloud geometry breaks 3D neighborhood continuity and weakens the model’s ability to capture spatial relations and directional variations. As illustrated in Figure.~\ref{fig1}, cross-modal attention collapses onto a few hotspots, many 3D tokens with clearly different positions and orientations receive nearly identical attention weights, and large regions of the scene are effectively ignored, resulting in pronounced spatial perception bias.

Although recent efforts have extended RoPE from 1D temporal position modeling to multimodal settings, these methods are primarily tailored to images or videos and do not align well with the characteristics of point-cloud tokens~\cite{wei2025videorope,ji2025ropetr,tu2024panopose,heo2024rotary}. While these approaches substantially improve the behavior of RoPE in their respective settings, they still treat positional indices as sequence- or grid-based and do not explicitly encode the 3D geometric structure of point-cloud tokens in 3D LVLMs. Motivated by the suboptimal behavior observed in Figure.~\ref{fig1}, we further analyze RoPE from an internal modeling perspective and attribute its key limitations in encoding point-cloud token positions to the following two factors:

\newenvironment{finding}[1]
{\par\smallskip\noindent\textbf{#1:}\ \ignorespaces}
{\par\smallskip}

\begin{finding}{Finding} \textit{(i) RoPE fails to preserve essential 3D spatial structure when modeling positions of 3D tokens. (ii) RoPE’s relative position term is direction-agnostic and cannot capture orientation changes.}
\end{finding}

\begin{itemize}
 \item We propose SoPE, a connector-level positional encoding for 3D LVLMs that remaps 3D tokens from a 1D raster index to a spherical tuple $(t, r, \theta, \phi)$, and redistributes RoPE frequency bands with multi-scale phase mixing so that temporal order, depth, and orientation are encoded in a geometry-consistent, direction-aware way.
 \item Plugged into SpatialLM as a drop-in replacement for vanilla RoPE, SpatialSoPE consistently improves 3D layout estimation and object detection across benchmarks and produces more balanced, geometry-aware cross-modal attention patterns.
\end{itemize}
To validate the practical applicability of our method, we performed an end-to-end deployment on a real device and reported measured metrics such as latency, frame rate, and power consumption.

\section{Related Work}

\subsection{3D Large Vision-Language Models}

Large multimodal models (LMMs), and in particular large vision-language models (LVLMs), have advanced rapidly, extending large language model (LLM) reasoning to visual and 3D inputs. A common paradigm projects images or 3D representations into the language embedding space so that visual features can be jointly processed with text tokens. 

A first line of 3D LVLMs~\cite{hong20233d,xu2024pointllm,chen2024ll3da,zhu2024llava,ma20253dsrbench,daxberger2025mm,zhou2025llava,zhu2025struct2d,huang2025zero,khattak2025good,chen2025idea23d,li2025lpl3d,liu2025reasongrounder,lubberstedt2025v3lma,wang2025dynamic,hu2024omnimedvqa,dong2025lumigen,jiang2024open,zemskova20253dgraphllm,chen2024spatialvlm,yu2025language,yan2024med,fan2025vlm,lyu2024mmscan,wang2025aigv} focuses on strengthening 3D visual representations. For example, ScanReason, PointLLM and LL3DA employ dedicated 3D encoders to directly model point-cloud geometry, while Video-3D LVLMs and Scene-LLM extract multi-view image features with 3D awareness; LLaVA-3D further injects point-cloud features into multi-view image tokens to produce 2D tokens enriched with 3D context. 

Another line of work advances 3D capability from data, architectural, and memory perspectives. SpatialLM~\cite{ma2025spatialllm} proposes a compound 3D-informed design of data, architectures and training to improve 3D spatial reasoning; Inst3D LMM~\cite{yu2025inst3d} achieves instance-level 2D–3D fusion and relation modeling via Multi-view Cross-Modal Fusion and 3D-ISR; 3DLLM-MEM~\cite{hu20253dllm} equips embodied 3D agents with dynamic working and episodic memory and introduces 3DMEM-BENCH to evaluate long-term spatiotemporal memory. LSceneLLM~\cite{zhi2025lscenellm} leverages LLM visual preferences~\cite{gao2024aligning} to guide a scene magnifier with adaptive attention, refining representations from sparse to fine-grained for cross-room scenes and introducing the XR-Scene benchmark.

\subsection{Position Encoding Schemes in LMMs}

Most 3D LMMs adopt the RoPE commonly used in LLMs: they flatten images into 1D tokens in raster order and apply complex phase rotations in the embedding space. This approach, however, breaks spatial neighborhood continuity, introduces end of sequence bias along the temporal dimension, and causes attention dilution as views and frames increase. To address these issues, various 2D image and video improvements have been proposed: HoPE~\cite{chen2024hope} selectively preserves high frequency components; CoMemo~\cite{liu2025comemo} and MiniCPM-V~\cite{yu2025minicpm} design RoPE variants to mitigate long range attenuation; CCA~\cite{zhao2025mca} and MCA~\cite{zhao2025mca} use heuristic index assignments to reduce hallucinations in long sequences; and VideoRoPE~\cite{wei2025videorope}, M‑RoPE~\cite{qwen2-vl} and decompose positional embeddings into 1D, 2D, and 3D components with hierarchical frequency distributions to improve spatiotemporal performance. New ideas for multi view and 3D sequences are also emerging: ComRoPE~\cite{yu2025comrope} uses a trainable exchangeable angle matrix to improve scaling and robustness; FoPE~\cite{hua2024fourier} restores spectral damage in the frequency domain to enhance long sequence generalization; DRoPE~\cite{zhao2025drope} adds Directionality to RoPE to handle heading periodicity; CR2PQ~\cite{zhangcr2pq} leverage two view 3D priors to improve matching and geometric reconstruction.And Circle-RoPE~\cite{wang2025circle} introduces a cone-like geometric projection to decouple cross-modal positional encodings.  Overall, despite a range of effective strategies, a systematic unified positional encoding scheme for multi view, long temporal 3D sequences remains lacking and warrants further study.
\section{Motivation and Analysis}

In this section, we provide an overview of the architecture of baseline 3D LVLMs and explain how RoPE is applied within them. Furthermore, through information flow analysis, we reveal the negative impact of RoPE on multimodal understanding, which forms a strong foundation for our proposed design.

\begin{figure}[t]
\centering
\includegraphics[page=1,width=0.5\textwidth]{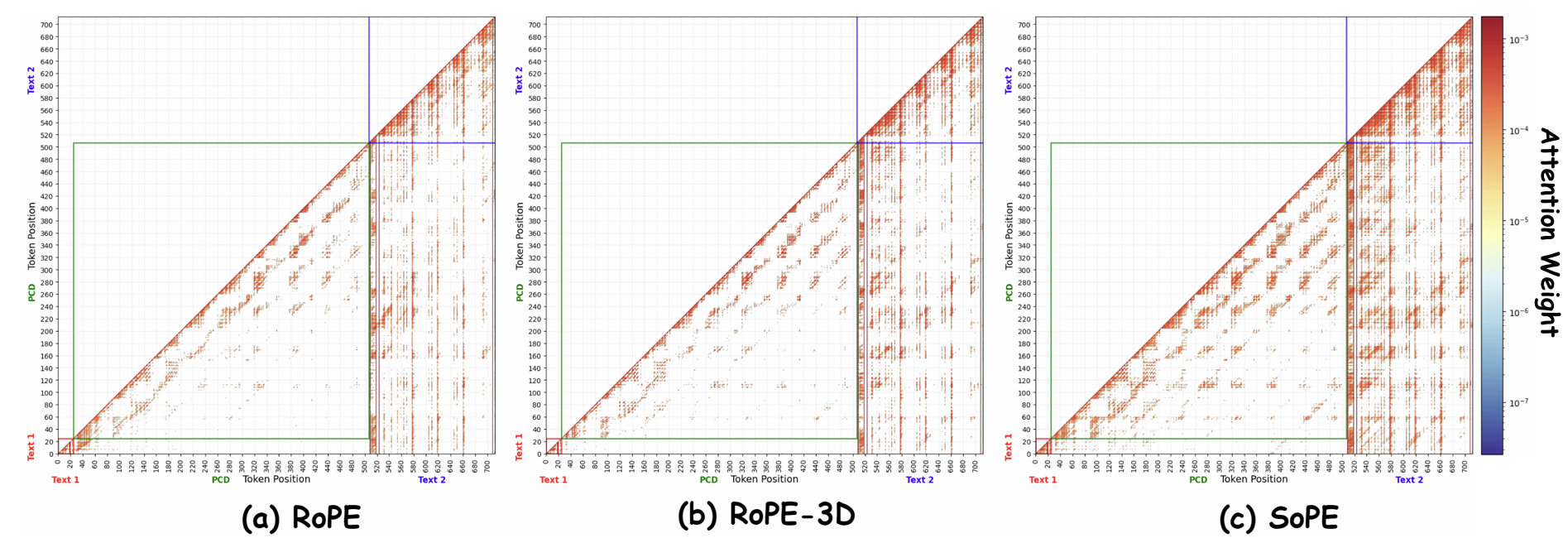} 
\caption{\textbf{Comparison of cross-modal attention patterns with RoPE, RoPE-3D, and SoPE}. (a) With RoPE, attention is predominantly intra-modal, failing to establish effective connections between the text and the point cloud. (b) RoPE-3D yields marginal improvements in cross-modal attention but remains limited. (c) The proposed SoPE successfully establishes stronger cross-modal attention links and substantially enhances information flow between the two modalities.}
\label{fig2}
\end{figure} 

\subsection{Preliminary}
\noindent
\textbf{3D LVLMs.} We adopt SpatialLM~\cite{SpatialLM} as our baseline 3D LVLM, denoted by $\mathcal{F}$. It is composed of three key components: a point cloud encoder $\mathcal{F}_p$, a large language model (LLM) $\mathcal{F}_t$, and a projector $f$ designed to map point cloud embeddings into the text embedding space. Given a point cloud input $I_p$ and a textual prompt input $I_t$, the model $\mathcal{F}$ encodes them into a multimodal sequence $\mathcal{S} = \{ \mathcal{S}_p , \mathcal{S}_t \}$, where $\mathcal{S}_p = f(\mathcal{F}p(I_p)) = \{ w_m \}_{m=1}^{P}$ and $\mathcal{S}_t = \mathcal{F}t(I_t) = \{ w_m \}_{m=1}^{T}$ represent point cloud and text tokens of lengths $V$ and $T$, respectively.

\noindent
\textbf{Rotary Position Encoding.}
As the 3D multimodal successor of LLMs, SpatialLM inherits RoPE to encode the
relative positional information of each token $w_m$. Specifically, given a query
token $Q_{t_{1}}$ at position $t_{1}$ and a key token $K_{t_{2}}$ at position
$t_{2}$, both with the same dimension $d$, RoPE divides the dimension $d$ into
$d/2$ pairs, with each pair assigned predefined sinusoidal function values
\[
\left\{ \theta'_i = 10000^{-2(i-1)/d} \right\}, \quad i \in \{ 1, 2, \dots, d/2 \}, 
\]
where $\theta'_i$\footnote{To avoid confusion with the polar angle $\theta$ used
later in spherical coordinates, we denote the RoPE base angles as $\theta'_i$.}
denote the RoPE base angles. Subsequently, the rotation matrix $r^{(i)}$ is
constructed as
\begin{equation}
r^{(i)} =
\begin{pmatrix}
  \cos(\theta'_i) & -\sin(\theta'_i) \\
  \sin(\theta'_i) &  \cos(\theta'_i) \\
\end{pmatrix}
\end{equation}
The relative rotation matrix $R_{\Delta t}$ is defined as
$R_{\Delta t} = \mathrm{diag}(\Delta t\, r^{(1)}, \dots, \Delta t\, r^{(d/2)})$,
where $\Delta t$ denotes the relative positional distance. The vanilla RoPE uses
$d = 128$ and encodes the relative positional relationship between $Q_{t_{1}}$ and
$K_{t_{2}}$ through the following relative matrix:
\begin{equation}
\resizebox{\columnwidth}{!}{
$\left(
\begin{array}{c}
Q^{(0)} \\[3pt]
Q^{(1)} \\[3pt]
\vdots \\[3pt]
Q^{(126)} \\[3pt]
Q^{(127)}
\end{array}
\right)^{\!\top}
\!
\left(
\begin{array}{ccccc}
\cos \theta'_0 \Delta t & -\sin \theta'_0 \Delta t & \cdots & 0 & 0 \\[3pt]
\sin \theta'_0 \Delta t &  \cos \theta'_0 \Delta t & \cdots & 0 & 0 \\[3pt]
\vdots & \vdots & \ddots & \vdots & \vdots \\[3pt]
0 & 0 & \cdots & \cos \theta'_{63} \Delta t & -\sin \theta'_{63} \Delta t \\[3pt]
0 & 0 & \cdots & \sin \theta'_{63} \Delta t &  \cos \theta'_{63} \Delta t
\end{array}
\right)
\!
\left(
\begin{array}{c}
K^{(0)} \\[3pt]
K^{(1)} \\[3pt]
\vdots \\[3pt]
K^{(126)} \\[3pt]
K^{(127)}
\end{array}
\right)$
}
\label{eq:2}
\end{equation}

\subsection{Spatial Perception Bias}
We conduct an in-depth analysis of the spatial perception bias introduced by RoPE through information flow visualization, revealing its negative impact on 3D multimodal understanding. Finally, we clarify the limitations of RoPE’s positional modeling mechanism and attribute the spatial perception bias to deficiencies in positional indexing and relative distance computation.

\noindent
\textbf{Information flow for Spatial Perception Bias.}
We visualize the information flow from the point cloud to the textual prompt to examine the spatial perception bias caused by RoPE. As shown in Figure.~\ref{fig2}, we observe that most 3D point cloud tokens exhibit sparse information flow. This indicates an uneven perception of visual features in 3D space by textual tokens during cross-modal interaction, where only a subset of point cloud tokens receive high attention. We refer to this phenomenon as spatial perception bias. Beyond the global hotspot pattern, we further observe that small objects and structural boundaries in large indoor scenes are often suppressed: their corresponding 3D tokens receive almost uniformly low attention even when they are geometrically and semantically salient. This suggests that a single positional scale in RoPE is not sufficiently sensitive to small angular variations and local geometric changes. In the following, we attribute the causes of spatial perception bias to two key limitations from the perspective of position-dependent modeling.

\noindent
\textbf{Hard embedding of 3D point cloud tokens.} The 3D point cloud tokens are flattened into a one-dimensional sequence and assigned positional indices following a raster-scan order. Although this approach is straightforward, it neglects the inherent difference between the spatial continuity of point cloud features and the discreteness of textual features, thereby disrupting the locality of three-dimensional spatial structures. In particular, spatially adjacent point cloud tokens are assigned non-adjacent positional indices.

\noindent
\textbf{Temporal-dependent relative position computation}. In RoPE, the relative distance between the query token $Q_{t_{1}}$ and the key token $K_{t_{2}}$ is computed as follows:
\begin{equation}
\Delta t = t_{1} - t_{2}.
\label{t}
\end{equation}
This distance computation can only capture the temporal variation of 3D tokens within the sequence. Standard RoPE is inherently unaware of the actual spatial position changes of tokens, a problem that recent variants like 3D-RoPE attempt to address by mapping visual tokens into spatial coordinates. However, these methods still fail to perceive crucial directional variations in visual representations, which is a key attribute of spatial awareness. Our proposed method directly targets this remaining gap.

The two aforementioned limitations form a solid foundation for our design. To address them, we propose a Spherical Coordinate-Based Positional Embedding to enhance the model’s perception of global spatial information. Inspired by prior 2D LVLMs that assign two-dimensional coordinates to image tokens to provide spatial cues, we further explore a simple RoPE variant that assigns three-dimensional coordinates to 3D tokens, referred to as RoPE-3D, which serves as a comparative baseline for our method.

\section{Method}

\begin{figure*}[h]
\centering
\includegraphics[width=0.9\textwidth]{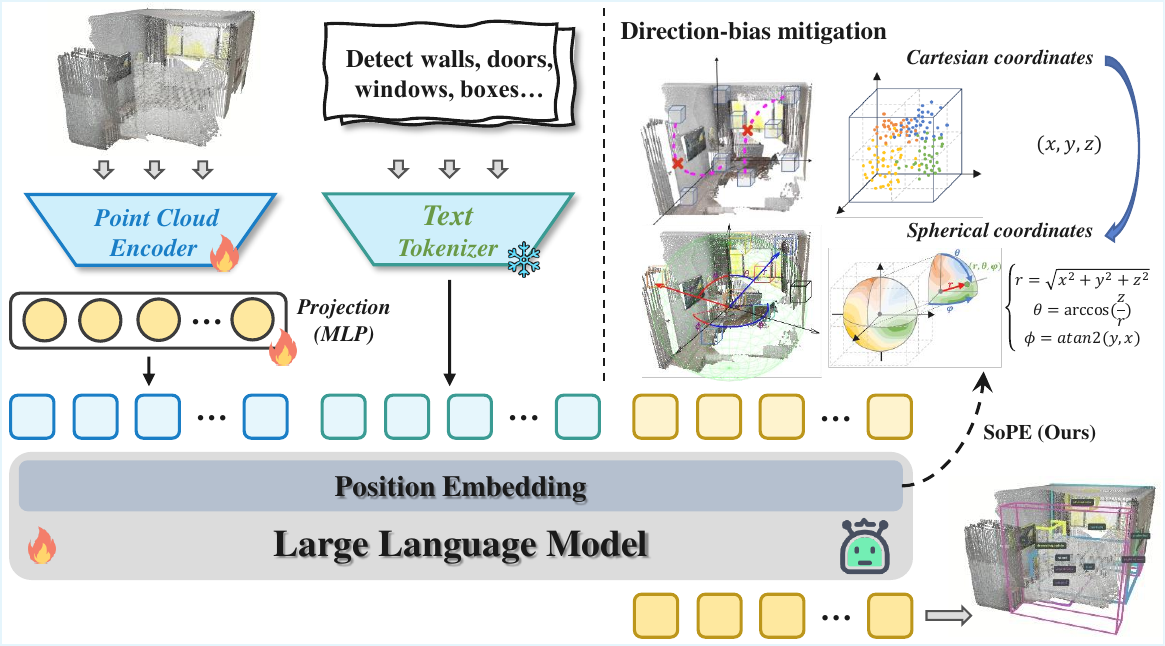} 
\caption{SoPE: Spherical Coordinate-Based Positional Embedding for Multimodal 3D Scene Understanding. 
The figure illustrates the overall SpatialSoPE framework, where SoPE is integrated into SpatialLM. 
By transforming coordinates from Cartesian to spherical, SpatialSoPE mitigates spatial directional bias 
and enables more robust 3D object understanding and localization.}
\label{fig3}
\end{figure*}

We propose Spherical Coordinate–Based Positional Embedding (SoPE), a positional encoding variant tailored to 3D point-cloud tokens. As illustrated in Figure~\ref{fig3}, SoPE reparameterizes point-cloud tokens from Cartesian coordinates into a spherical coordinate system, enhancing the model’s sensitivity to both spatial positions and directional variations. SoPE comprises three key components: spherical coordinate positional projection, multi-dimensional frequency allocation, and multi-scale frequency encoding.


\subsection{Spherical Coordinate Positional Projection}\label{4.1}

Spherical coordinate positional projection, decouples 3D spatial tokens from the 1D textual sequence by replacing the flattened raster index with a geometry-aware tuple $(t, r, \theta, \phi)$.This process comprises two key steps:

(i) Positional Index Reallocation: We extract the Cartesian coordinates $(x,y,z)$ of point cloud tokens to represent their intrinsic 3D spatial positions, while preserving the original RoPE-assigned index $t$ to maintain their temporal ordering within the sequence. Each token is thereby assigned a reallocated four-dimensional positional index $(t,x,y,z)$, integrating both sequential and spatial information for unified positional encoding.

(ii) Spherical Coordinate Mapping: We project the Cartesian positional index components into a spherical coordinate space, where the three spherical components radius $r$, polar angle $\theta$, and azimuthal angle $\phi$ are computed as follows:
\begin{equation}
\begin{aligned}
r &= \sqrt{x^2 + y^2 + z^2}, \\
\theta &= \arccos\left( \frac{z}{r} \right), \\
\phi &= \operatorname{atan2}(y, x).
\end{aligned}
\end{equation}

We then combine the temporal index $t$ with the three spherical components to obtain the final composite four-dimensional positional index $(t,r, \theta , \phi)$. Finally, we effectively embed both spatial location and angular orientation of point cloud tokens into the positional modeling scheme. Building upon this enriched representation, we further generalize RoPE’s original relative position formulation (Equation~\ref{t}) by decomposing the displacement across all four components:
\begin{equation}
\begin{aligned}
\Delta t = t_{1} - t_{2}&, &\Delta r = r_{1} - r_{2}, \\
\Delta \theta = \theta_{1} - \theta_{2}&, &\Delta \phi = \phi_{1} - \phi_{2}.
\end{aligned}
\end{equation}
This extended formulation enables the model to capture spatial and directional variations when computing relative positional relationships between point cloud tokens.


\subsection{Multi-dimensional Frequency Allocation}


In Sec. \ref{4.1}, we extend the temporal positional index to include four components: temporal position $t$, radius $r$, polar angle $\theta$, and azimuthal angle $\phi$. In this work we split the frequency spectrum into four proportional parts corresponding to these components, with an allocation ratio  $t : r : \theta : \phi = 24 : 2 : 3 : 3$. To determine this design we ran extensive ablation experiments comparing many alternative allocations and band-orderings; the results guided our choice of both the per-component bandwidths and the subband ordering used in the final model. Concretely, we map the three spherical-coordinate components $r, \theta, \phi$ to the front higher-frequency subbands and place the temporal component $t$ in the later lower-frequency subbands. Intuitively, assigning the spherical components to higher frequencies helps the model capture fine-grained spatial and angular variations and directional cues, while allocating a larger low-frequency block to the temporal component preserves slower, long-range temporal dynamics and continuity. This allocation achieved the best trade-off between spatial and angular resolution and temporal coherence in our ablation studies. The final relative rotation matrix is formulated as follows:

\begin{equation}
\label{eq:sope_combined_clean}
\resizebox{\columnwidth}{!}{
$
\begin{aligned}
& \left(
\begin{array}{c}
Q^{(0)} \\[3pt]
Q^{(1)} \\[3pt]
Q^{(2)} \\[3pt]
Q^{(3)} \\[3pt]
\vdots \\[3pt]
Q^{(6)} \\[3pt]
Q^{(7)}
\end{array}
\right)^{\!\top}
\!
\left(
\begin{array}{cccccc}
\cos\theta'_{0}\Delta r & -\sin\theta'_{0}\Delta r & 0 & \cdots & 0 & 0 \\[3pt]
\sin\theta'_{0}\Delta r &  \cos\theta'_{0}\Delta r & 0 & \cdots & 0 & 0 \\[3pt]
0 & 0 & \cos\theta'_{1}\Delta r & -\sin\theta'_{1}\Delta r & \cdots & 0 \\[3pt]
0 & 0 & \sin\theta'_{1}\Delta r &  \cos\theta'_{1}\Delta r & \cdots & 0 \\[3pt]
\vdots & \vdots & \vdots & \vdots & \ddots & \vdots \\[3pt]
0 & 0 & \cdots & 0 & \cos\theta'_{3}\Delta r & -\sin\theta'_{3}\Delta r \\[3pt]
0 & 0 & \cdots & 0 & \sin\theta'_{3}\Delta r &  \cos\theta'_{3}\Delta r
\end{array}
\right)
\!
\left(
\begin{array}{c}
K^{(0)} \\[3pt]
K^{(1)} \\[3pt]
K^{(2)} \\[3pt]
K^{(3)} \\[3pt]
\vdots \\[3pt]
K^{(6)} \\[3pt]
K^{(7)}
\end{array}
\right) \\
+ & \left(
\begin{array}{c}
Q^{(8)} \\[3pt]
Q^{(9)} \\[3pt]
Q^{(10)} \\[3pt]
Q^{(11)} \\[3pt]
\vdots \\[3pt]
Q^{(18)} \\[3pt]
Q^{(19)}
\end{array}
\right)^{\!\top}
\!
\left(
\begin{array}{cccccc}
\cos\theta'_{4}\Delta \theta & -\sin\theta'_{4}\Delta \theta & 0 & \cdots & 0 & 0 \\[3pt]
\sin\theta'_{4}\Delta \theta &  \cos\theta'_{4}\Delta \theta & 0 & \cdots & 0 & 0 \\[3pt]
0 & 0 & \cos\theta'_{5}\Delta \theta & -\sin\theta'_{5}\Delta \theta & \cdots & 0 \\[3pt]
0 & 0 & \sin\theta'_{5}\Delta \theta &  \cos\theta'_{5}\Delta \theta & \cdots & 0 \\[3pt]
\vdots & \vdots & \vdots & \vdots & \ddots & \vdots \\[3pt]
0 & 0 & \cdots & 0 & \cos\theta'_{9}\Delta \theta & -\sin\theta'_{9}\Delta \theta \\[3pt]
0 & 0 & \cdots & 0 & \sin\theta'_{9}\Delta \theta &  \cos\theta'_{9}\Delta \theta
\end{array}
\right)
\!
\left(
\begin{array}{c}
K^{(8)} \\[3pt]
K^{(9)} \\[3pt]
K^{(10)} \\[3pt]
K^{(11)} \\[3pt]
\vdots \\[3pt]
K^{(18)} \\[3pt]
K^{(19)}
\end{array}
\right) \\
+ & \left(
\begin{array}{c}
Q^{(20)} \\[3pt]
Q^{(21)} \\[3pt]
Q^{(22)} \\[3pt]
Q^{(23)} \\[3pt]
\vdots \\[3pt]
Q^{(30)} \\[3pt]
Q^{(31)}
\end{array}
\right)^{\!\top}
\!
\left(
\begin{array}{cccccc}
\cos\theta'_{10}\Delta \phi & -\sin\theta'_{10}\Delta \phi & 0 & \cdots & 0 & 0 \\[3pt]
\sin\theta'_{10}\Delta \phi &  \cos\theta'_{10}\Delta \phi & 0 & \cdots & 0 & 0 \\[3pt]
0 & 0 & \cos\theta'_{11}\Delta \phi & -\sin\theta'_{11}\Delta \phi & \cdots & 0 \\[3pt]
0 & 0 & \sin\theta'_{11}\Delta \phi &  \cos\theta'_{11}\Delta \phi & \cdots & 0 \\[3pt]
\vdots & \vdots & \vdots & \vdots & \ddots & \vdots \\[3pt]
0 & 0 & \cdots & 0 & \cos\theta'_{15}\Delta \phi & -\sin\theta'_{15}\Delta \phi \\[3pt]
0 & 0 & \cdots & 0 & \sin\theta'_{15}\Delta \phi &  \cos\theta'_{15}\Delta \phi
\end{array}
\right)
\!
\left(
\begin{array}{c}
K^{(20)} \\[3pt]
K^{(21)} \\[3pt]
K^{(22)} \\[3pt]
K^{(23)} \\[3pt]
\vdots \\[3pt]
K^{(30)} \\[3pt]
K^{(31)}
\end{array}
\right) \\
+ & \left(
\begin{array}{c}
Q^{(32)} \\[3pt]
Q^{(33)} \\[3pt]
Q^{(34)} \\[3pt]
Q^{(35)} \\[3pt]
\vdots \\[3pt]
Q^{(126)} \\[3pt]
Q^{(127)}
\end{array}
\right)^{\!\top}
\!
\left(
\begin{array}{cccccc}
\cos\theta'_{16}\Delta t & -\sin\theta'_{16}\Delta t & 0 & \cdots & 0 & 0 \\[3pt]
\sin\theta'_{16}\Delta t &  \cos\theta'_{16}\Delta t & 0 & \cdots & 0 & 0 \\[3pt]
0 & 0 & \cos\theta'_{17}\Delta t & -\sin\theta'_{17}\Delta t & \cdots & 0 \\[3pt]
0 & 0 & \sin\theta'_{17}\Delta t &  \cos\theta'_{17}\Delta t & \cdots & 0 \\[3pt]
\vdots & \vdots & \vdots & \vdots & \ddots & \vdots \\[3pt]
0 & 0 & \cdots & 0 & \cos\theta'_{63}\Delta t & -\sin\theta'_{63}\Delta t \\[3pt]
0 & 0 & \cdots & 0 & \sin\theta'_{63}\Delta t &  \cos\theta'_{63}\Delta t
\end{array}
\right)
\!
\left(
\begin{array}{c}
K^{(32)} \\[3pt]
K^{(33)} \\[3pt]
K^{(34)} \\[3pt]
K^{(35)} \\[3pt]
\vdots \\[3pt]
K^{(126)} \\[3pt]
K^{(127)}
\end{array}
\right)
\end{aligned}
$
}
\end{equation}

The matrices in Equations (6) are finally integrated into the overall rotation matrix.

\subsection{Multi-scale Frequency Mixing Strategy}
Single-scale positional encodings, even with spherical coordinates, struggle to capture both fine-grained geometry and large-scale architectural layouts in indoor 3D scenes, which motivates a multi-scale treatment of positional cues. To this end, we introduce a multi-scale frequency mixing strategy at the positional phase level. For each temporal or spherical coordinate component $u \in \{t, r, \theta, \phi\}$, we regard $u$ as a fixed geometric quantity and define different \emph{scales} as deterministic transforms applied to the same $u$ before RoPE phase computation. Concretely, we construct three complementary coordinate transforms $g^{\text{lin}}(u)$, $g^{\text{log}}(u)$, and $g^{\text{per}}(u)$, corresponding to a linear scale that preserves absolute positional precision, a log-compressed scale that emphasizes local neighborhood structure, and a periodic scale that captures global patterns and long-range dependencies.

In practice, multi-scale fusion is implemented directly at the RoPE phase level. For each frequency band $k$ assigned to component $u$, the final phase used in the $k$-th rotation block is:
\begin{equation}
\varphi_k(u) = \frac{1}{3}\Big(
\omega_k^{\text{lin}} g^{\text{lin}}(u) +
\omega_k^{\text{log}} g^{\text{log}}(u) +
\omega_k^{\text{per}} g^{\text{per}}(u)
\Big),
\end{equation}
where $\omega_k^{\cdot}$ denote the RoPE base frequencies at each scale. We use fixed uniform weights across the three scales without introducing additional learnable parameters, keeping the design lightweight while still yielding clear gains in our ablation studies. By mixing phases across scales in this manner, SoPE can simultaneously encode precise locations, local context, and global 3D structure, leading to more balanced attention patterns and stronger cross-modal links.
\section{Experiment}
\subsection{Experimental Setup}\label{1}

Following SpatialLM~\cite{SpatialLM}, we adopt Sonata~\cite{Sonata} as the point cloud encoder and train it jointly with the rest of the model. Qwen2.5-0.5B~\cite{Qwen2.5} as the LLM, and a two-layer MLP~\cite{cai2024large,fan2024stockmixer} to bridge the two modules. All experiments are conducted in a single-stage training setup using four NVIDIA H20 GPUs. We evaluate our method on three 3D scene benchmarks for layout estimation and 3D object detection. We adopt IoU$_{2D}$ as the metric for layout estimation and IoU$_{3D}$ for 3D object detection~\cite{SpatialLM}. For all ablation studies, we keep the experimental settings and test splits identical to the original SpatialLM setup to ensure fair comparisons. SpatialSoPE denotes SpatialLM equipped with the proposed SoPE positional encoding. 

\begin{table}[t!]
\centering
\footnotesize 
\caption{ \textbf{Integration of SoPE enables state-of-the-art layout estimation}. Our results indicate that the SpatialSoPE module, when fine‑tuned (ft), is the main factor enabling us to outperform prior methods such as RoomFormer and SceneScript.} 
\label{tab1}
\resizebox{\columnwidth}{!}{%
    \begin{tabular}{ccc} 
    \toprule
    \multirow{2}{*}{Methods} & \multicolumn{2}{c}{F1 - Structured3D} \\
    \cmidrule(lr){2-3}
 & IoU$_{2D}$ @0.25$\uparrow$ & IoU$_{2D}$@0.5$\uparrow$ \\ 
    \midrule
    RoomFormer~\cite{yue2023connecting} & 70.4 & 67.2 \\ 
    SceneScript~\cite{avetisyan2024scenescript} & 83.1 & 80.8 \\ 
    \midrule
    SpatialLM (ft. Structured3D)~\cite{SpatialLM} & 32.8 & 17.9 \\ 
    SpatialSoPE (ft. Structured3D) & 33.5 & 29.5 \\ 
    SpatialLM (ft. SpatialLM Dataset $\rightarrow$ Structured3D) & 86.5 & 84.6 \\ 
    SpatialSoPE (ft. SpatialLM Dataset) $\rightarrow$ Structured3D) & \textbf{88.7} & \textbf{86.2} \\ 
    \bottomrule
    \end{tabular}%
}
\end{table}

\begin{table}[t!]
\centering
\caption{\textbf{3D object detection results on the ARKitScenes dataset}. SpatialSoPE consistently 
outperforms SpatialLM and other 3D LVLM baselines.}
\label{tab2}
\resizebox{\columnwidth}{!}{
\begin{tabular}{ccc}
\toprule
\multirow{2}{*}{Methods} & \multicolumn{2}{c}{F1 - ARKitScenes} \\
\cmidrule(lr){2-3}
& IoU$_{3D}$ @0.25$\uparrow$ & IoU$_{3D}$ @0.50$\uparrow$ \\
\midrule
VoteNet~\cite{ding2019votenet}      & 53.9 & 45.4 \\
H3DNet~\cite{zhang2020h3dnet}     & 55.7 & 46.3 \\
NeRF-Det~\cite{xu2023nerf}      & 60.3 & 34.7 \\
UniDet3D~\cite{kolodiazhnyi2025unidet3d}      & 62.8 & 48.3 \\
SpatialLM~\cite{SpatialLM}      & 63.9 & 60.7 \\
SpatialSoPE & \textbf{66.1} & \textbf{63.2} \\
\bottomrule
\end{tabular}
}
\end{table}

\begin{table}[t!]
\centering
\caption{\textbf{Comparison between SpatialSoPE and SpatialLM on the SpatialLM Dataset and Structured3D}. SpatialSoPE consistently outperforms the SpatialLM baseline across all metrics on both datasets.}
\label{tab3}
\resizebox{\columnwidth}{!}{
\begin{tabular}{ccccc}
\toprule
\multirow{2}{*}{Methods} & \multicolumn{2}{c}{F1 - SpatialLM Dataset} & \multicolumn{2}{c}{F1 - Structured3D} \\
\cmidrule(lr){2-3} \cmidrule(lr){4-5}
& IoU$_{3D}$ @0.25$\uparrow$ & IoU$_{3D}$ @0.50$\uparrow$ & IoU$_{3D}$ @0.25$\uparrow$ & IoU$_{3D}$ @0.50$\uparrow$ \\
\midrule
SpatialLM~\cite{SpatialLM}      & 69.7 & 62.0 & 31.8 & 17.6 \\
SpatialSoPE    & \textbf{71.4} & \textbf{63.4} & \textbf{33.3} & \textbf{17.9} \\
\bottomrule
\end{tabular}
}
\end{table}

\subsection{Experimental comparison results.}
\textbf{Results of Layout Estimation.}
The layout estimation task tests a model’s 3D spatial perception by predicting architectural elements in indoor scenes. On Structured3D, SpatialSoPE reaches IoU$_{2D}$@0.25 and IoU$_{2D}$@0.5 of 88.7 and 86.2, achieving gains of +2.2 and +1.6 over SpatialLM (Table~\ref{tab1}). We observe consistent improvements under both training-from-scratch and pretrain-then-finetune configurations, indicating that the geometry- and direction-aware positional cues of SoPE are robust across training protocols and data scales.

\begin{figure}[t]
\centering
\includegraphics[page=1,width=0.95\columnwidth]{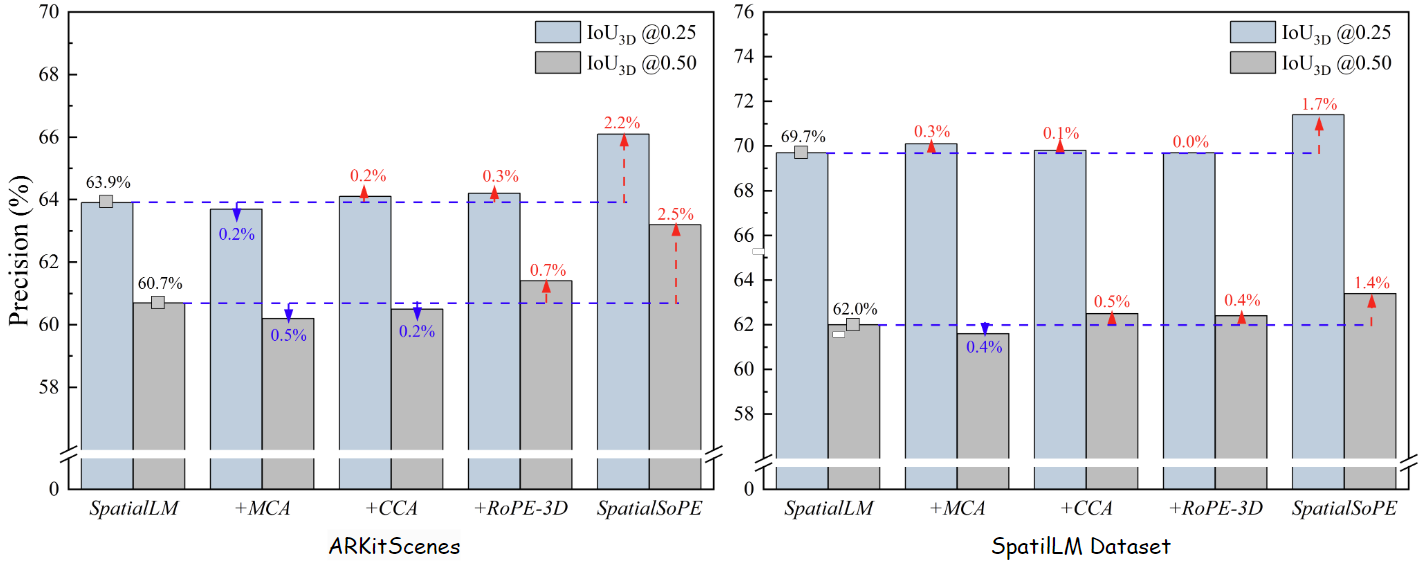}
\caption{\textbf{Effectiveness of the SoPE module demonstrated through component analysis.} The figure compares various components added to the SpatialLM, highlighting the superior contribution of SoPE to model precision. Results are shown for (a) the ARKitScenes and (b) the SpatialLM datasets.}
\label{fig4}
\end{figure}

\noindent
\textbf{Results of 3D Object Detection.}
We further evaluate SpatialSoPE on three 3D object detection benchmarks: ARKitScenes~\cite{baruch2021arkitscenes}, the SpatialLM Dataset, and Structured3D~\cite{zheng2020structured3d}. Integrating SoPE yields consistent improvements on all datasets (Tables~\ref{tab2} and~\ref{tab3}): on ARKitScenes, IoU$_{3D}$ F1@0.25/@0.50 improves by +2.2/+2.5; on the SpatialLM Dataset by +1.7/+1.4; and on Structured3D by +1.5/+0.3. These gains, particularly under looser IoU thresholds, support that SoPE’s hierarchical, geometry-aware encoding better captures positional information and directional variations in point-cloud tokens, leading to stronger 3D spatial understanding.

\begin{figure*}[t!]
\centering
\includegraphics[page=1,width=0.92\textwidth]{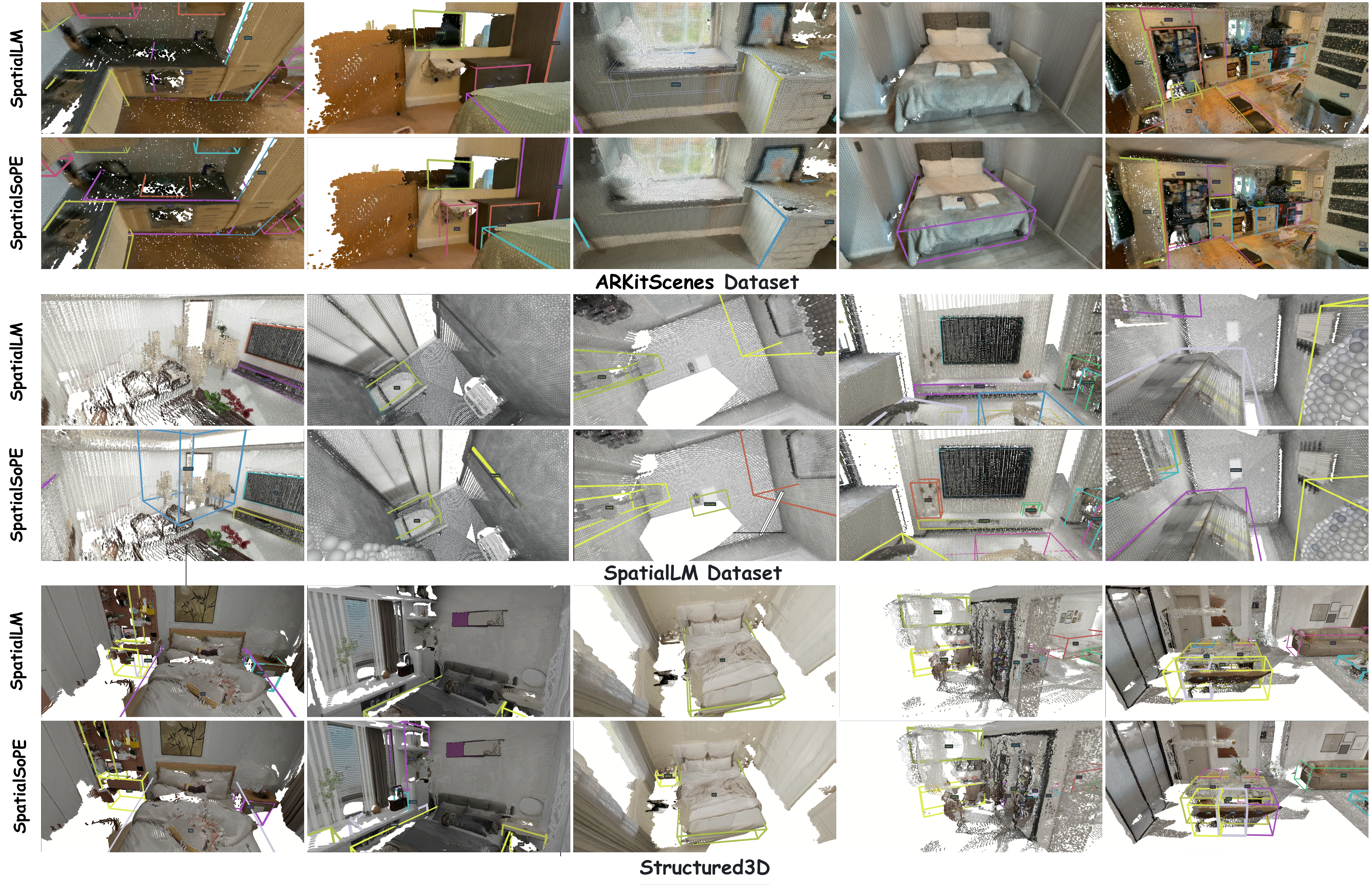} 
\caption{\textbf{Visualization of improvements in 3D object detection with SpatialSoPE}. Compared to SpatialLM, our method demonstrates significantly improved localization and sizing of 3D bounding boxes across various scenes.}
\label{fig5}
\end{figure*} 

\noindent
\textbf{Comparison of different positional encoding methods.} As shown in Table~\ref{tab4}, we evaluate three representative RoPE-based multimodal variants CCA, MCA, and RoPE-3D against our method to highlight the advantages of SpatialSoPE. CCA and MCA first project visual token indices onto a 2D plane and apply heuristic reassignment rules: CCA arranges indices in a communication-square pattern that increases from the boundary inward, while MCA computes indices directly from 2D spatial coordinates. In our reproduction, we extend these strategies to the 3D setting by projecting point-cloud tokens onto a 2D plane and applying analogous index reassignments. However, as the table shows, both methods yield suboptimal performance, likely because collapsing 3D structures into 2D discards crucial spatial locality and directional information. RoPE-3D, in contrast, shows that direct spatial encoding is a more effective paradigm, improving the IoU$_{3D}$@0.50 F1 score from 60.7 to 61.4 by explicitly incorporating 3D coordinates. 

Compared with the above methods, the proposed SpatialSoPE integrates a SoPE module that projects point cloud tokens into a 3D spherical coordinate space, jointly encoding spatial positions and directional angles. This geometry-aware positional encoding better matches the inherent 3D structure of point clouds, enhancing spatial awareness and improving IoU$_{3D}$ F1 to 66.1 and 63.2 at thresholds 0.25 and 0.50, respectively. A qualitative comparison is provided in Figure.~\ref{fig4}.

\begin{table}[t!]
\centering
\caption{\textbf{Analysis of Component Contributions}. We demonstrate the impact of each key module by comparing the full SpatialSoPE model against the baseline and its intermediate variants on the ARKitScenes and SpatialLM datasets.}
\label{tab4}
\resizebox{\columnwidth}{!}{
\begin{tabular}{ccccc}
\toprule
\multirow{2}{*}{Methods} & \multicolumn{2}{c}{F1 - ARKitScenes} & \multicolumn{2}{c}{F1 - SpatialLM Dataset}\\
\cmidrule(lr){2-3} \cmidrule(lr){4-5}
& IoU$_{3D}$ @0.25$\uparrow$ & IoU$_{3D}$ @0.50$\uparrow$ & IoU$_{3D}$ @0.25$\uparrow$ & IoU$_{3D}$ @0.50$\uparrow$ \\
\midrule
SpatialLM~\cite{SpatialLM}      & 63.9 & 60.7 & 69.7 & 62.0 \\
+MCA~\cite{zhao2025mca} & 63.7 & 60.2 & 70.1 & 61.6 \\
+CCA~\cite{cca-llava} & 64.1 & 60.5 & 69.8 & 62.5 \\
+RoPE-3D       & 64.2 & 61.4 & 69.7 & 62.4 \\
\midrule
SpatialSoPE    & \textbf{66.1} & \textbf{63.2} & \textbf{71.4} & \textbf{63.4} \\
\bottomrule
\end{tabular}
}
\end{table}

\subsection{Ablation Studies}
\begin{table}[t]
\centering
\caption{\textbf{Ablation study for our proposed SpatialSoPE method}. We validate our parameter choices by comparing against several variants on the ARKitScenes dataset.}
\label{tab5}
\footnotesize
\setlength{\tabcolsep}{3.5pt}
\begin{tabular}{cccc} 
\toprule
\multirow{2}{*}{Methods}  & \multirow{2}{*}{$t:r:\theta:\phi$} & \multicolumn{2}{c}{F1 - ARKitScenes} \\
\cmidrule(lr){3-4}
& & IoU$_{3D}$ @0.25$\uparrow$ & IoU$_{3D}$ @0.50$\uparrow$ \\
\midrule
Angular-Biased & (8:6:9:9) & 65.5 & 62.7 \\
Uniform & (1:1:1:1) & 63.0 & 59.0 \\
Temporal-Biased & (5:1:1:1) & 65.0 & 62.7 \\
\midrule
SpatialSoPE & (24:2:3:3) & \textbf{66.1} & \textbf{63.2} \\
\bottomrule
\end{tabular}
\end{table}

\begin{table}[t!]
\centering
\caption{\textbf{Effect of Multi-scale Frequency Mixing Strategy for RoPE-3D and SpatialSoPE}. We compare RoPE-3D and SpatialSoPE with and without the proposed Multi-scale Frequency Mixing Strategy, reporting F1 scores on ARKitScenes and the SpatialLM dataset.}
\label{tab6}
\resizebox{\columnwidth}{!}{
\begin{tabular}{ccccccc}
\toprule
\multirow{3}{*}{Method} &\multirow{3}{*}{Multi-scale} & \multicolumn{2}{c}{F1 - ARKitScenes} & \multicolumn{2}{c}{F1 - SpatialLM Dataset} & \\
\cmidrule(lr){3-4} \cmidrule(lr){5-6}
& & \multirow{1}{*}{IoU$_{3D}$ @0.25$\uparrow$} & \multirow{1}{*}{IoU$_{3D}$ @0.50$\uparrow$} & \multirow{1}{*}{IoU$_{3D}$ @0.25$\uparrow$} & \multirow{1}{*}{IoU$_{3D}$ @0.50$\uparrow$} & \\
\midrule
\multirow{2}{*}{SpatialLM+RoPE-3D}
& $\times$ & 64.2 & \textbf{61.7} & 69.4 & 62.3 \\
& $\checkmark$ & 64.8 & 62.1 & 70.3 & 62.9 \\
\midrule
\multirow{2}{*}{SpatialSoPE}
& $\times$ & \textbf{65.4} & 61.4 & \textbf{71.0} & \textbf{62.5} \\
& $\checkmark$ & \textbf{66.1} & \textbf{63.2} & \textbf{71.4} & \textbf{63.4} \\
\bottomrule
\end{tabular}
}
\end{table}

\textbf{Comparison of Frequency Allocation Schemes}. Under identical settings, we perform a controlled study that varies only the frequency allocation strategy in SpatialSoPE. As shown in Table~\ref{tab5}, our allocation consistently outperforms all alternatives, suggesting that overemphasizing temporal frequencies harms geometric and orientation modeling, while overemphasizing angular/spatial frequencies undermines sequence-level stability. By reserving low-frequency capacity for long-range temporal coherence and assigning high-frequency bandwidth to spherical components, SpatialSoPE more accurately separates local from global 3D structure.

\noindent
\textbf{Comparison of Multi-scale Frequency Mixing Strategy}. As shown in Table~\ref{tab6}, we further evaluate the Multi-scale Frequency Mixing Strategy on both SpatialLM+RoPE-3D and SpatialSoPE. For SpatialLM+RoPE-3D, although $(x,y,z)$ are encoded in Cartesian coordinates, multi-scale mixing brings only modest gains and mainly acts as a smoothing operation. In contrast, the same design yields clearly larger improvements for SpatialSoPE, where coarse scales preserve global layout and temporal stability, and fine scales emphasize small objects, boundaries, and orientation differences. Built on top of spherical reparameterization and tailored frequency allocation, the multi-scale strategy allows SpatialSoPE to better disentangle local and global 3D structure, leading to more stable detection and more accurate spatial perception.

\subsection{Case Study}
As illustrated in Figure~\ref{fig5}, we present a qualitative comparison of 3D object detection performance across three different point cloud sources. The figure horizontally displays five viewpoints for each scene to dissect the effect of SpatialSoPE. A direct visual comparison shows that the model equipped with SoPE substantially reduces false detections, particularly for objects that are small or geometrically intricate, thereby improving overall robustness. In addition, SpatialSoPE further strengthens the model’s sensitivity to orientation and angular cues, leading to more consistent detection results across multiple views of the same scene. Taken together, these qualitative visualizations, along with the quantitative evaluations presented above, provide strong evidence that SpatialSoPE significantly enhances the model’s spatial perception and cross-view understanding in 3D environments. 

\subsection{Real-World Validation}

\begin{figure}[t]
    \centering
    \includegraphics[width=0.85\columnwidth, height=0.45\columnwidth]{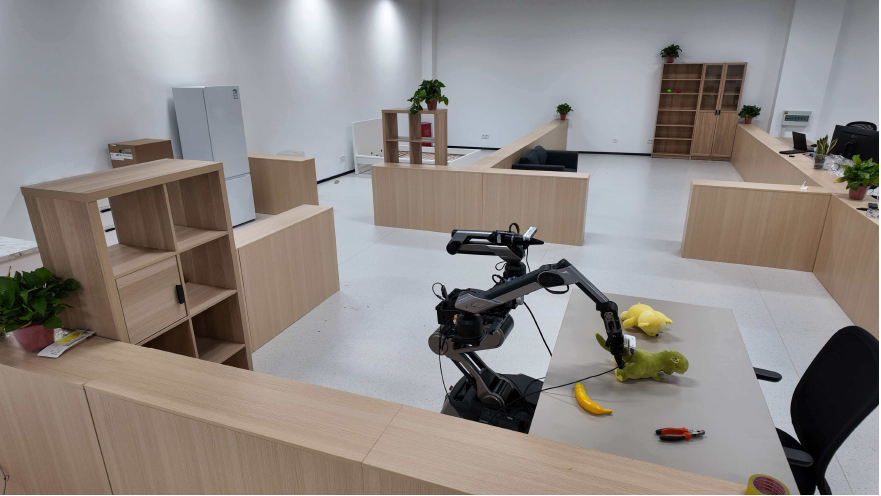}
    \caption{The figure illustrates the real-world experimental environment and the physical robot used in our study. Equipped with SpatialSoPE, the robot is capable of understanding the entire scene and executing exploration and object-transport tasks according to human instructions.}
    \label{fig:real}
    \vspace{-2.0em}
\end{figure}

SpatialSoPE’s strong spatial understanding of indoor environments has the potential to significantly advance embodied intelligence. To demonstrate its practical utility, we integrate SpatialSoPE into a complete robotic system and conduct real-world experiments, as illustrated in Fig.~\ref{fig:real}. First, we reconstruct and render a point cloud of the environment using MASt3R-SLAM~\cite{murai2025mast3r}. Next, SpatialSoPE processes the reconstructed scene to produce semantic labels, bounding boxes, and other object-level information. This information is then converted into a textual scene graph, from which we generate navigation target points—following a design similar to Work~\cite{habitatrearrangechallenge2022, gu2022multi, makhal2018reuleaux}—based on the bounding boxes of each piece of furniture for global path planning. Finally, inspired by Work~\cite{liu2025coherent, yu2024mhrc}, we employ LLMs as the robot’s high-level task planner, encapsulating navigation, grasping, placing, and other behaviors into atomic actions. Built upon SpatialSoPE’s scene understanding, this enables general and efficient task planning for the robot. More details are provided in the supplementary material.
\section{Conclusion}
In this work, we proposed SoPE, a spherical orientation positional encoding tailored for 3D LVLMs that reparameterizes tokens in the ($t : r : \theta : \phi$) space, redistributes RoPE frequencies across temporal and angular dimensions, and performs multi-scale phase mixing to better capture 3D geometry and direction. Instantiated as a drop-in replacement for RoPE in SpatialLM, the resulting SpatialSoPE model achieves consistent gains on 3D layout estimation and 3D object detection benchmarks, while controlled ablations verify the contribution of the spherical reparameterization, frequency allocation strategy, and multi-scale phase mixing individually. Qualitative analyses further show that SpatialSoPE reduces false detections on small and geometrically intricate objects and yields more consistent predictions across multiple viewpoints of the same scene, indicating improved spatial perception and cross-view reasoning. Finally, we demonstrate that SpatialSoPE can be deployed in a real-world robotic system, providing rich scene-understanding signals for robot planning and validating its feasibility and superiority.

{
    \small
    \bibliographystyle{ieeenat_fullname}
    \bibliography{main}
}
\clearpage
\setcounter{page}{1}
\maketitlesupplementary

\section{Real-World Validation}
\label{sec:real}

\subsection{Settings}

\begin{figure}[t]
    \centering
    \includegraphics[width=\columnwidth]{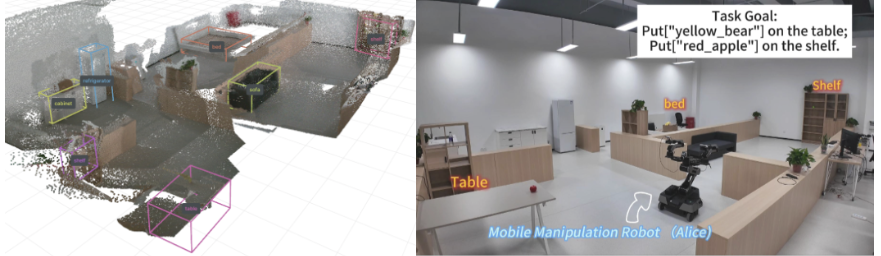}
    \caption{Detection visualization and task environments of SpatialSoPE in real-world reconstructed scenes.}
    \label{fig:sm}
\end{figure}

In this section, we present a detailed account of the real-robot experimental validation process. Fig.~\ref{fig:sm} illustrates the reconstructed scene point cloud, the model’s visual outputs, and the corresponding real-world setup. We build a simulated indoor home environment containing a table, sofa, bookshelf, refrigerator, and other common furniture. The experimental platform is the Galaxea R1 Lite mobile dual-arm robot, equipped with a wrist-mounted RealSense D435i, a RealSense D455, and a Mid-360 LiDAR. We design distinct atomic action sets for the robot to encapsulate its physical capabilities, including:

\begin{enumerate}
    \item \texttt{[navigate] to <stand\_pose\_id> of <object>};
    \item \texttt{[open] <container>};
    \item \texttt{[pick] up <object>};
    \item \texttt{[place] <object> on/into <platform>};
    \item \texttt{[move] <delta\_x> and <delta\_y>};
\end{enumerate}

Following the methodologies of prior work~\cite{liu2025coherent, yu2024mhrc}, we employ LLM-based role-playing to guide task planning through a Chain-of-Thought (CoT) framework. The robot selects and executes one action at a time from a predefined list. After each action, it receives feedback and, using this information together with its execution history, autonomously determines the next subtask or adjusts its plan. This multi-turn process enables effective coordination for long-horizon tasks. The detailed prompt design is provided in Subsection~\ref{subsec:prompts}.

In the real-robot system, for manipulation tasks, we use AnyGrasp to predict grasp points and apply Grounded SAM to filter candidates based on semantic targets. The resulting grasp poses are executed via inverse-kinematics planning. For navigation tasks, we adopt a SLAM pipeline for mapping and localization, with A* as the global planner and DWA as the local planner.

We further integrate the multimodal \textit{GPT-4o} model to obtain scene understanding and action-level feedback in real-world environments. After each action, the manipulator’s wrist camera captures the current scene to assess execution success and overall task progress. For example, once the mobile robot reaches a new piece of furniture, \textit{GPT-4o} identifies the objects on it to support exploration in unknown environments. Similarly, after each picking or placing operation, \textit{GPT-4o} verifies whether the action was executed correctly. Additional feedback, such as pose estimation or inverse-kinematics results, can also be incorporated into the evaluation process.

\subsection{Prompts}
\label{subsec:prompts}
\promptblock{\small
\commenttext{==== System Prompt ====} \\
\textbf{Role:}
\begin{enumerate}
\item You are an intelligent robot named \$\{name\}, configured with a wheeled chassis and a single manipulator arm. 
\item You possess the ability to navigate across the ground and perform manipulation tasks, including transporting various objects and opening hinged objects.
\end{enumerate}
\textbf{Skills:}
\begin{enumerate}
\item navigate(obj, pose); open(obj);
\item pick(obj); place(obj, loc); move(dx, dy);
\end{enumerate}
\textbf{Task Objective and Context:}
\begin{enumerate}
\item The overall task is: \$\{target\_task\}.
\item Ingredients are scattered in an unknown indoor environment. The scene graph shows furniture locations but not their contents. Based on task goal, objects must be placed.
\end{enumerate}
\textbf{Principles:}
\begin{enumerate}
\item Efficiently explore and navigate all locations in the scene graph without repetition.
\item Transport task-related items promptly.
\item Track task progress and adjust targets timely.
\item If grasp fails, try other stand poses or adjust base position.
\item Focus on completing the task without unrelated actions.
\end{enumerate}
\textbf{Output Response Format:}
\begin{enumerate}
\item Thoughts: think step by step to analyze the problem;
\item Contents: choose and execute only one action from the action functions above.
\end{enumerate}
\textbf{CoT:}
Let's think step by step! \\
\textbf{Examples:}
The following examples are provided for reference in decision-making. The related content involved has nothing to do with the actual task: [...] \\
\commenttext{==== User Prompt ====} \\
\textbf{Scene Graph:} \\
drawer: (pos: [...], ori: [...], state: close, stand: [...]), \\
cabinet: (...), sofa: (...), ... (It is generated by SpatialSoPE)\\
\textbf{Robot Status:} \\
Current robot states: pos: [...], ori: [...]. The gripper is empty. \\
\textbf{Feedback History:} \\
The historical feedbacks, from oldest to newest, are as follows: [... Robot successfully reached the target book case-stand pose 0. [`book 0', `book 1'] are found on the book case ...] \\
\textbf{Action History:} \\
The historical actions, from oldest to newest, are as follows: [... navigate(book case, stand pose 0) ...]
}

%


\end{document}